\documentclass{article}
\usepackage{ijcai26}

\usepackage{times}
\usepackage{soul}
\usepackage{url}
\usepackage[hidelinks]{hyperref}
\usepackage[utf8]{inputenc}
\usepackage[small]{caption}
\usepackage{graphicx}
\usepackage{amsmath}
\usepackage{amsthm}
\usepackage{booktabs}
\usepackage{algorithm}
\usepackage{algorithmic}
\usepackage[switch]{lineno}

\usepackage{multirow}
\usepackage{makecell}
\usepackage{colortbl}
\usepackage{xcolor}
\usepackage[table]{xcolor}
\definecolor{lightblue}{RGB}{235, 245, 255}
\definecolor{lightgray}{gray}{0.9}
\usepackage[most]{tcolorbox}
\usepackage{subcaption}
\usepackage{url}

\title{Rewarding How Models Think Pedagogically:\\Integrating Pedagogical Reasoning and Thinking Rewards for LLMs in Education}

\author{
Unggi Lee$^{1\dagger}$ \and
Jiyeong Bae$^2$\and
Jaehyeon Park$^3$\and
Haeun Park$^4$ \and
Taejun Park$^3$ \\
Younghoon Jeon$^5$ \and
Sungmin	Cho$^2$ \and
Junbo Koh$^3$ \and
Yeil Jeong$^{6\dagger}$ \And
Gyeonggeon Lee$^{7\dagger}$ \\
\affiliations
$^1$Chosun University 
$^2$Korea University
$^3$Seoul National University \\
$^4$Korea Institute for Curriculum and Evaluation 
$^5$Upstage \\
$^6$Indiana University Bloomington
$^7$Nanyang Technological University \\
\textbf{Corresponding Authors (†, contact emails):} codingchild@korea.ac.kr, gyeonggeon.lee@nie.edu.sg
}

\begin{document}

\maketitle

\begin{abstract}

Large language models (LLMs) are increasingly deployed as intelligent tutoring systems, yet research on optimizing LLMs specifically for educational contexts remains limited. Recent works have proposed reinforcement learning approaches for training LLM tutors, but these methods focus solely on optimizing visible responses while neglecting the model's internal thinking process. We introduce PedagogicalRL-Thinking, a framework that extends pedagogical alignment to reasoning LLMs in education through two novel approaches: (1) Pedagogical Reasoning Prompting, which guides internal reasoning using domain-specific educational theory rather than generic instructions; and (2) Thinking Reward, which explicitly evaluates and reinforces the pedagogical quality of the model's reasoning traces. Our experiments reveal that domain-specific, theory-grounded prompting outperforms generic prompting, and that Thinking Reward is most effective when combined with pedagogical prompting. Furthermore, models trained only on mathematics tutoring dialogues show improved performance on educational benchmarks not seen during training, while preserving the base model's factual knowledge. Our quantitative and qualitative analyses reveal that pedagogical thinking reward produces systematic reasoning trace changes, with increased pedagogical reasoning and more structured instructional decision-making in the tutor's thinking process.

\end{abstract}
\section{Introduction}

Large language models (LLMs) are increasingly deployed as intelligent tutoring systems, offering scalable one-on-one educational support across diverse subject domains \cite{kasneci2023chatgpt,roll2016evolution}. However, a fundamental challenge persists: LLM tutors often provide direct answers rather than guiding students through the problem-solving process, undermining the pedagogical value of the interaction \cite{macneil2023experiences,tack2023bea}. This tendency reflects a misalignment between the model's optimization objective, typically next-token prediction (e.g., answer), and the educational goal of fostering student understanding through scaffolded guidance (e.g., process).

Recent work has begun addressing this challenge through reinforcement learning (RL) approaches. Notably, PedagogicalRL \cite{jurenka2024towards} formulates tutoring as an RL problem with rewards for solution correctness and pedagogical quality \cite{dinucujianu2025pedagogy,lee2025pedagogyr1}. However, such approaches still focus exclusively on the \textit{visible response}, the text presented to students, while ignoring the model's internal \textit{thinking process}. This limitation is particularly notable given the emergence of reasoning-specialized LLMs \cite{deepseek2025r1} that explicitly generate intermediate reasoning traces before producing final responses.

This paper addresses a central research question: Can applying pedagogical rewards to the thinking process, in addition to visible responses, improve LLM tutor performance? We hypothesize that rewarding pedagogically appropriate reasoning, not just pedagogically appropriate outputs, will produce tutors that more consistently guide students without revealing answers.

\begin{figure*}[t]
    \centering
    \includegraphics[width=1\linewidth]{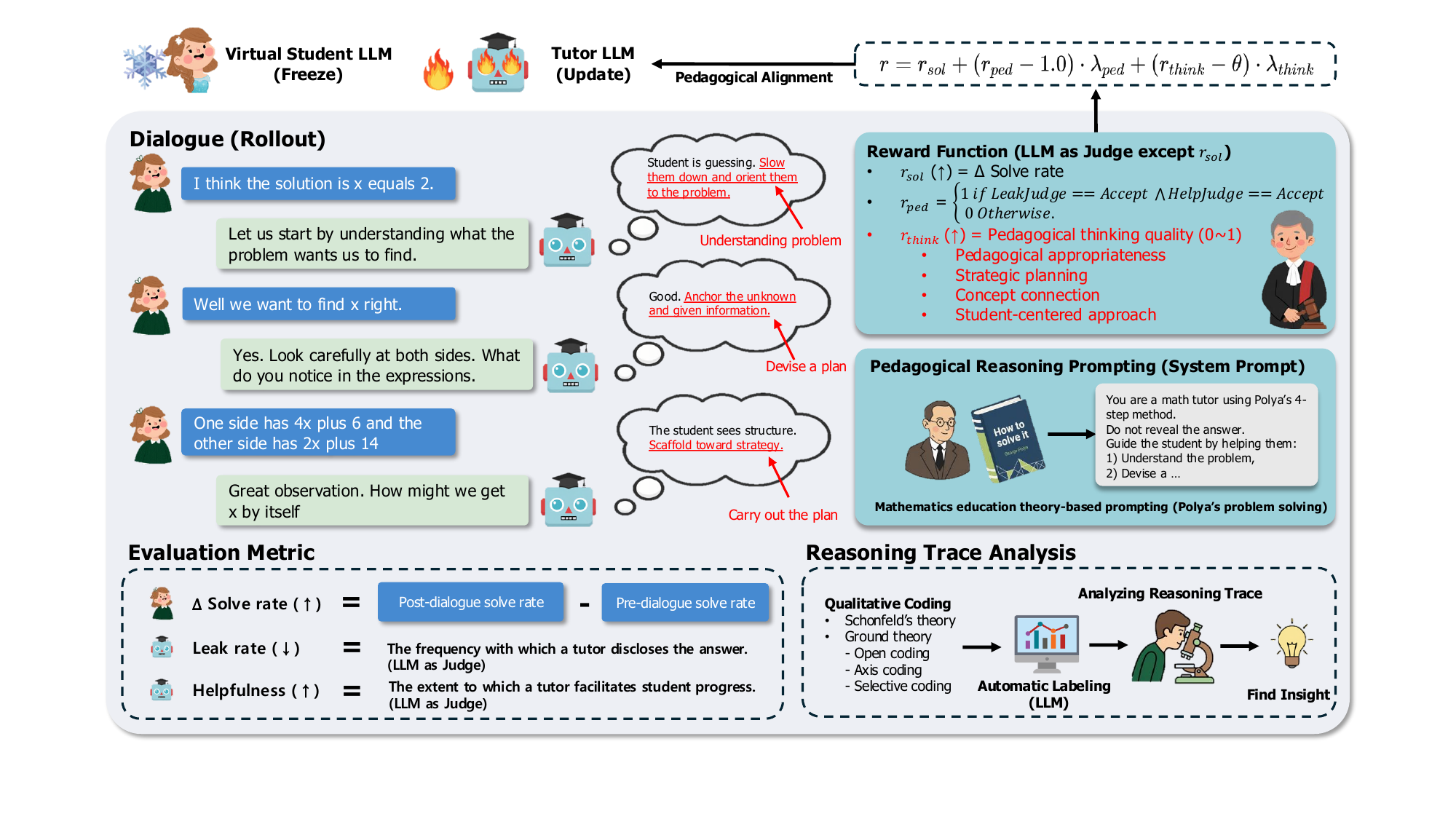}
    \caption{Overview of PedagogicalRL-Thinking. A frozen virtual student LLM interacts with an updatable tutor LLM through multi-turn dialogues. Our framework integrates two components. (1) Pedagogical Reasoning Prompting guides the tutor's internal reasoning using Polya's four-step problem-solving method. (2) Thinking Reward explicitly evaluates the pedagogical quality of reasoning traces ($r_{think}$). The composite reward function combines solve rate improvement ($r_{sol}$), pedagogical appropriateness ($r_{ped}$), and thinking quality ($r_{think}$). We evaluate tutors on $\Delta$ solve rate, leak rate, and helpfulness, and analyze reasoning traces through qualitative coding.}
    \label{fig:overview}
\end{figure*}

We introduce PedagogicalRL-Thinking\footnote{\url{https://anonymous.4open.science/r/pedagogical_reasoning_submission-2787/README.md}}, a framework that extends pedagogical reinforcement learning to the thinking phase of reasoning LLMs (Figure~\ref{fig:overview}). Our approach integrates two complementary components: (1) Pedagogical Reasoning Prompting, which replaces generic tutoring instructions with domain-specific prompts grounded in mathematics education theory (Polya's four-step problem-solving framework \cite{polya1945solve}); and (2) Thinking Reward, an LLM-based evaluation of the pedagogical quality of reasoning traces, providing explicit training signal for the thinking process.

\subsection{Contributions}

We evaluate PedagogicalRL-Thinking across five experimental conditions that systematically ablate thinking capability, Pedagogical Reasoning Prompting, and Thinking Reward. Our main contributions are:
\begin{itemize}
    \item We present the first application of pedagogical rewards to the internal thinking process of reasoning LLMs, demonstrating that rewarding \textit{how} models think, not just what they output, improves educational outcomes.
    \item We demonstrate that domain-specific, theory-grounded prompting outperforms general prompting for educational tasks. Our Pedagogical Reasoning Prompting, based on Polya's problem-solving method, significantly improves both thinking quality and visible response quality compared to generic tutoring instructions.
    \item We show that dialogue-based RL training yields out-of-distribution generalization to educational benchmarks not seen during training, while preserving the base model's factual knowledge.
    \item We provide a comprehensive qualitative analysis framework using an 82-code educational codebook, revealing systematic differences in tutoring behavior across conditions with statistical validation.
\end{itemize}

\section{Method}

\subsection{Problem Formulation}

We formulate LLM tutoring as a Markov Decision Process (MDP) where the tutor interacts with a student over multiple dialogue turns. At each turn $t$, the state $s_t$ comprises the dialogue history including the math problem, all previous student messages, and tutor responses. The action $a_t$ is the tutor's response, consisting of a thinking component $a_t^{\text{think}}$ enclosed in \texttt{<think>} tags and a visible component $a_t^{\text{visible}}$ shown to the student. The reward $r_t$ is a composite signal evaluating solution correctness, pedagogical quality, and thinking quality. The objective is to learn a policy $\pi_\theta$ that maximizes expected cumulative reward while maintaining pedagogically appropriate behavior.

\subsection{Base Models}

We employ two base models depending on the experimental condition. For all thinking-enabled conditions, we use DeepSeek-R1-0528-Qwen3-8B (hereafter referred to as Qwen3-8B), a reasoning-specialized model that naturally generates thinking traces within \texttt{<think>} tags. We selected this model because it demonstrated superior thinking quality and controllable reasoning budget in our preliminary experiments. For the NoThink baseline condition, we use Qwen-2.5-7B, a standard instruction-tuned model without explicit thinking capability.

\subsection{Pedagogical Reasoning Prompting}

Pedagogical Reasoning Prompting is grounded in Polya's four-step problem-solving method \cite{polya1945solve}, instructing the tutor to guide students through: (1) understanding the problem, (2) devising a plan, (3) carrying out the plan, and (4) looking back. For example: ``\textit{Guide the student by helping them understand the problem, devise a plan, carry out the plan step by step, and verify whether the answer makes sense.}'' The full prompts are provided in Appendix~\ref{appendix:prompts}.

\begin{table*}[t]
\centering
\scriptsize
\caption{
Performance comparison on BigMath tutoring task. $\Delta$ Solve = Delta Solve Rate (improvement in student problem-solving after tutoring), Leak = Leak Rate (proportion of answer-revealing responses), Helpful = Helpful Rate (proportion of educationally helpful responses). $\uparrow$ indicates higher is better; $\downarrow$ indicates lower is better. Ped.\ Think Reward achieves the best performance across all metrics, demonstrating that combining pedagogical prompting with thinking reward yields optimal tutoring behavior.
}
\begin{tabular*}{\textwidth}{@{\extracolsep{\fill}}llcccccc}
\toprule
\multirow{2}{*}{\textbf{Models}} & \multirow{2}{*}{\textbf{Base Model}} & \multicolumn{3}{c}{\textbf{Configuration}} & \multicolumn{3}{c}{\textbf{Performance}} \\
\cmidrule(lr){3-5} \cmidrule(lr){6-8}
& & Think & Prompt Style & Reward & $\Delta$ Solve $\uparrow$ & Leak $\downarrow$ & Helpful $\uparrow$ \\
\midrule
\rowcolor{lightgray}
\multicolumn{8}{c}{\textbf{Frontier Models (Zero-shot)}} \\
\midrule
GPT-5.2 & -- & Think & Normal & None & 0.290 & \textbf{0.000} & 0.500 \\
GPT-5.2 & -- & Think & Ped. & None & 0.340 & \textbf{0.000} & 0.440 \\
Claude-4-Opus & -- & Think & Normal & None & 0.330 & 0.120 & 0.620 \\
Claude-4-Opus & -- & Think & Ped. & None & \underline{0.350} & \underline{0.090} & 0.760 \\
Gemini-3-Pro & -- & Think & Normal & None & 0.010 & 0.390 & 0.610 \\
Gemini-3-Pro & -- & Think & Ped. & None & 0.080 & 0.400 & 0.500 \\
DeepSeek-V3.2 & -- & Think & Normal & None & 0.260 & \underline{0.090} & \underline{0.770} \\
DeepSeek-V3.2 & -- & Think & Ped. & None & \textbf{0.390} & 0.110 & \textbf{0.820} \\
\midrule
\rowcolor{lightblue}
\multicolumn{8}{c}{\textbf{RL-Trained Models}} \\
\midrule
NoThink & Qwen2.5-7B & NoThink & Normal & None & 0.120 & 0.300 & 0.180 \\
Think NoReward & Qwen3-8B & Think & Normal & Res. Only & 0.281 & \underline{0.180} & 0.730 \\
Think Reward & Qwen3-8B & Think & Normal & Res. + Think & \underline{0.284} & 0.182 & 0.764 \\
Ped. Think NoReward & Qwen3-8B & Think & Ped. & Res. Only & 0.275 & 0.214 & \underline{0.766} \\
Ped. Think Reward (Ours) & Qwen3-8B & Think & Ped. & Res. + Think & \textbf{0.294} & \textbf{0.172} & \textbf{0.776} \\
\bottomrule
\end{tabular*}
\label{tab:main_results}
\end{table*}

\subsection{Reward Design}

Our reward function consists of three components, each addressing a distinct aspect of tutoring quality.

For solution correctness ($r_{\text{sol}}$), following \cite{jurenka2024towards}, we evaluate whether the tutoring dialogue improves the student's problem-solving ability. After the dialogue concludes, the student model attempts the problem $K=8$ times, and we compute the average success rate as $r_{\text{sol}}$.

For pedagogical quality ($r_{\text{ped}}$), we employ two LLM-based judges (GPT-4o-mini) to assess pedagogical appropriateness. The Leak Judge determines whether the tutor revealed the answer or solution steps that the student should discover independently. The Help Judge assesses whether the tutor's response provides meaningful educational guidance. We set $r_{\text{ped}} = 1.0$ if both judges accept the response, and $r_{\text{ped}} = 0.0$ otherwise.

For thinking quality ($r_{\text{think}}$), our key contribution is explicitly rewarding the thinking process. An LLM judge (GPT-4o-mini) evaluates the content within \texttt{<think>} tags for pedagogical appropriateness of reasoning, consideration of student understanding, and metacognitive awareness in planning responses. The score $r_{\text{think}} \in [0, 1]$ reflects the overall pedagogical quality of the thinking trace.

The composite reward combines all components:
\begin{equation}
r = r_{\text{sol}} + (r_{\text{ped}} - 1.0) \cdot \lambda_{\text{ped}} + (r_{\text{think}} - \theta) \cdot \lambda_{\text{think}}
\end{equation}
where $\lambda_{\text{ped}} = 0.75$ penalizes pedagogically inappropriate responses, $\lambda_{\text{think}} = 0.3$ weights the thinking reward, and $\theta = 0.6$ sets the thinking quality threshold.

\subsection{Experimental Conditions}

We design five conditions to ablate the contributions of thinking, pedagogical prompting, and thinking rewards. \textit{NoThink} (no thinking capability) serves as a baseline using Qwen-2.5-7B. \textit{Think NoReward} and \textit{Think Reward} (thinking enabled with and without thinking reward) use DeepSeek-R1-Qwen3-8B with standard prompting. \textit{Ped.\ Think NoReward} and \textit{Ped.\ Think Reward} (pedagogical prompting with thinking) investigate the combined effect of pedagogical instruction and thinking rewards.

\subsection{Training Algorithm}

We use Group Relative Policy Optimization (GRPO) \cite{shao2024deepseekmath} implemented in the veRL framework. GRPO computes advantages by comparing outcomes within groups of sampled responses, avoiding the need for a separate value model. For each training step, we sample a batch of 16 problems, generating 8 complete dialogue rollouts per problem for a total of 128 rollouts per batch. Each dialogue is limited to a maximum of 16 turns. Training uses 4 GPUs with hyperparameters detailed in Appendix~\ref{appendix:training}.

\subsection{Evaluation}

We evaluate tutor performance using three metrics. \textit{Delta Solve Rate} ($\Delta$ Solve) measures the improvement in student problem-solving success after tutoring, computed as the difference between post-dialogue and pre-dialogue solve rates with $K=8$ attempts. \textit{Leak Rate} quantifies the proportion of tutor responses that reveal the answer or key solution steps, assessed by GPT-4o-mini. \textit{Helpful Rate} measures the proportion of responses judged as providing meaningful educational guidance, also assessed by GPT-4o-mini.

To assess whether dialogue-based RL affects general knowledge, we additionally evaluate on the Well-balanced Educational Benchmark (WBEB), which covers subject knowledge, pedagogical knowledge, essay scoring, and teacher decision-making. WBEB tasks are evaluated using multiple-choice question accuracy.

\begin{table*}[t]
\centering
\scriptsize
\caption{Generalization performance on WBEB educational benchmark. SK = Subject Knowledge, PK = Pedagogical Knowledge, AES = Automated Essay Scoring, DM = Decision Making. All metrics are accuracy (\%). Results demonstrate that dialogue-based RL training preserves the base model's general educational knowledge.}
\label{tab:wbeb}
\begin{tabular*}{\textwidth}{@{\extracolsep{\fill}}llccccccc}
\toprule
\multirow{2}{*}{\textbf{Model}} & \multirow{2}{*}{\textbf{Base Model}} & \multicolumn{3}{c}{\textbf{Configuration}} & \multicolumn{4}{c}{\textbf{WBEB Performance}} \\
\cmidrule(lr){3-5} \cmidrule(lr){6-9}
& & Think & Prompt Style & Reward & SK $\uparrow$ & PK $\uparrow$ & AES $\uparrow$ & DM $\uparrow$ \\
\midrule
\rowcolor{lightgray}
\multicolumn{9}{c}{\textbf{Original Model}} \\
\midrule
Qwen3-8B & -- & Think & -- & None & 23.2 & 21.1 & 10.7 & 15.8 \\
\midrule
\rowcolor{lightblue}
\multicolumn{9}{c}{\textbf{RL-Trained Models}} \\
\midrule
Think NoReward & Qwen3-8B & Think & Normal & Res. Only & 21.9 (\textcolor{red}{-1.3}) & 27.6 (\textcolor{green!50!black}{+6.5}) & \underline{23.7} (\textcolor{green!50!black}{+13.0}) & \textbf{54.4} (\textcolor{green!50!black}{+38.6}) \\
Think Reward & Qwen3-8B & Think & Normal & Res. + Think & 21.7 (\textcolor{red}{-1.5}) & \textbf{29.5} (\textcolor{green!50!black}{+8.4}) & 23.6 (\textcolor{green!50!black}{+12.9}) & 53.4 (\textcolor{green!50!black}{+37.6}) \\
Ped.\ Think NoReward & Qwen3-8B & Think & Ped. & Res. Only & \underline{22.4} (\textcolor{red}{-0.8}) & \underline{29.0} (\textcolor{green!50!black}{+7.9}) & \underline{23.7} (\textcolor{green!50!black}{+13.0}) & 53.1 (\textcolor{green!50!black}{+37.3}) \\
Ped.\ Think Reward & Qwen3-8B & Think & Ped. & Res. + Think & \textbf{22.5} (\textcolor{red}{-0.7}) & 26.0 (\textcolor{green!50!black}{+4.9}) & \textbf{25.5} (\textcolor{green!50!black}{+14.8}) & \underline{53.8} (\textcolor{green!50!black}{+38.0}) \\
\bottomrule
\end{tabular*}
\end{table*}

\section{Experiments}

\subsection{Experimental Setup}

We use BigMath \cite{albalak2025bigmath} for training and evaluation, a large-scale mathematics problem dataset spanning arithmetic, algebra, and word problems. Following prior work \cite{jurenka2024towards,dinucujianu2025pedagogy}, we employ Llama-3.1-8B-Instruct as a student simulator that interacts with the tutor model during training and evaluation. The simulator is prompted to behave as a student learning mathematics.

For data selection, we filter problems based on student model solve rate following \cite{dinucujianu2025pedagogy}. We select problems where Llama-3.1-8B-Instruct achieves a solve rate between 1\% and 60\%, ensuring problems are neither trivially easy nor impossibly difficult for meaningful tutoring. This yields 10,000 problems for training and 500 problems for evaluation.

We evaluate tutor performance using three metrics: (1) Delta Solve Rate ($\Delta$ Solve), the improvement in student problem-solving success rate after tutoring compared to before, computed as the difference between post-dialogue and pre-dialogue solve rates with $K=8$ student attempts; (2) Leak Rate, the proportion of tutor responses that reveal the answer or key solution steps, assessed by GPT-4o-mini; and (3) Helpful Rate, the proportion of responses judged as providing meaningful educational guidance by GPT-4o-mini.

\subsection{Performance}

Table~\ref{tab:main_results} presents the performance results across all experimental conditions, including frontier models for comparison. Ped.\ Think Reward achieves the best performance among our trained models across all three metrics: highest Delta Solve Rate (0.294), lowest Leak Rate (0.172), and highest Helpful Rate (0.776).

The results reveal several key findings. First, enabling thinking dramatically improves performance: comparing NoThink to Think NoReward, we observe a +134\% improvement in Delta Solve Rate (0.120 $\to$ 0.281) and a +306\% improvement in Helpful Rate (0.180 $\to$ 0.730). Second, pedagogical prompting combined with thinking reward produces the best results, with Ped.\ Think Reward outperforming all ablated versions. Third, the thinking reward is most effective when combined with pedagogical prompting: comparing Ped.\ Think NoReward to Ped.\ Think Reward shows a 6.9\% improvement in Delta Solve Rate and a 19.6\% reduction in Leak Rate.

\subsection{Effect of Dialogue-Based RL on General Knowledge}

To assess whether dialogue-based reinforcement learning affects the base model's general knowledge, we evaluate our models on the Well-balanced Educational Benchmark (WBEB), a comprehensive benchmark covering subject knowledge, pedagogical knowledge, essay scoring, and teacher decision-making. We originally included knowledge tracing (KT) in our evaluation; however, all models achieved AUC scores near 0.5, indicating no meaningful discriminative ability on this task, so we excluded it from our analysis.

Table~\ref{tab:wbeb} presents the WBEB evaluation results. Notably, our RL training used only mathematics tutoring dialogues and did not include any WBEB-related data. Despite this, all RL-trained models show substantial improvements in pedagogical knowledge (+4.9 to +8.4\%p), essay scoring (+12.9 to +14.8\%p), and decision making (+37.3 to +38.6\%p). Among all conditions, Ped.\ Think Reward achieves the most balanced performance: it minimizes subject knowledge degradation (-0.7\%p) while attaining the highest AES score (25.5\%). Although subject knowledge slightly decreases across all conditions, the degradation remains within 1.5 percentage points, indicating that the base model's factual knowledge is largely preserved during dialogue-based RL training. This out-of-distribution improvement aligns with recent findings that RL training can enhance performance on domains not directly targeted during training \cite{deepseek2025r1,liu2025guru}.

\section{Analysis}

To understand why Ped.\ Think Reward performs best, we conduct comprehensive quantitative and qualitative analyses of tutor responses.

\subsection{Response Characteristics}

\begin{figure*}[t]
\centering
\begin{subfigure}[b]{0.32\textwidth}
\centering
\includegraphics[width=\textwidth]{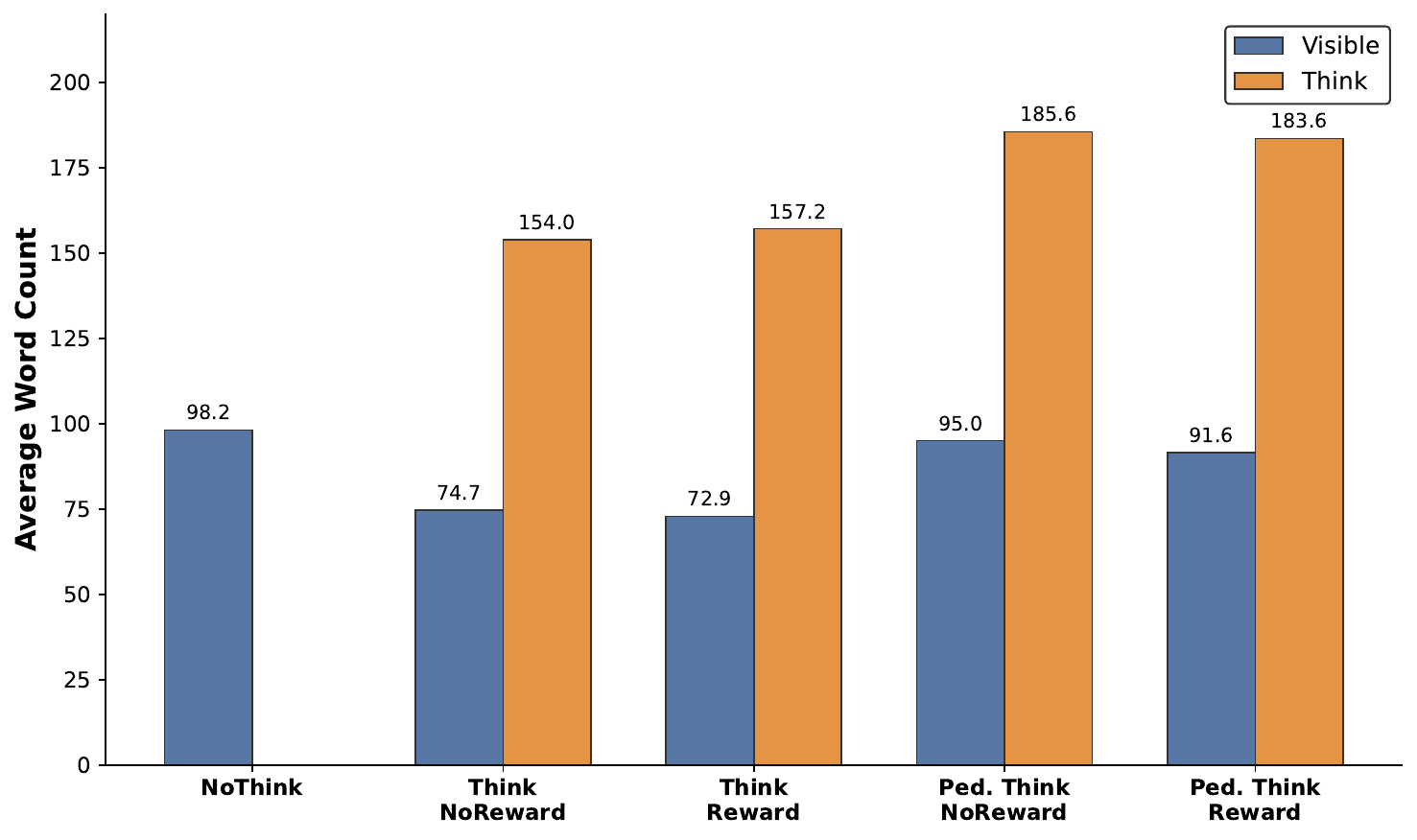}
\end{subfigure}
\hfill
\begin{subfigure}[b]{0.32\textwidth}
\centering
\includegraphics[width=\textwidth]{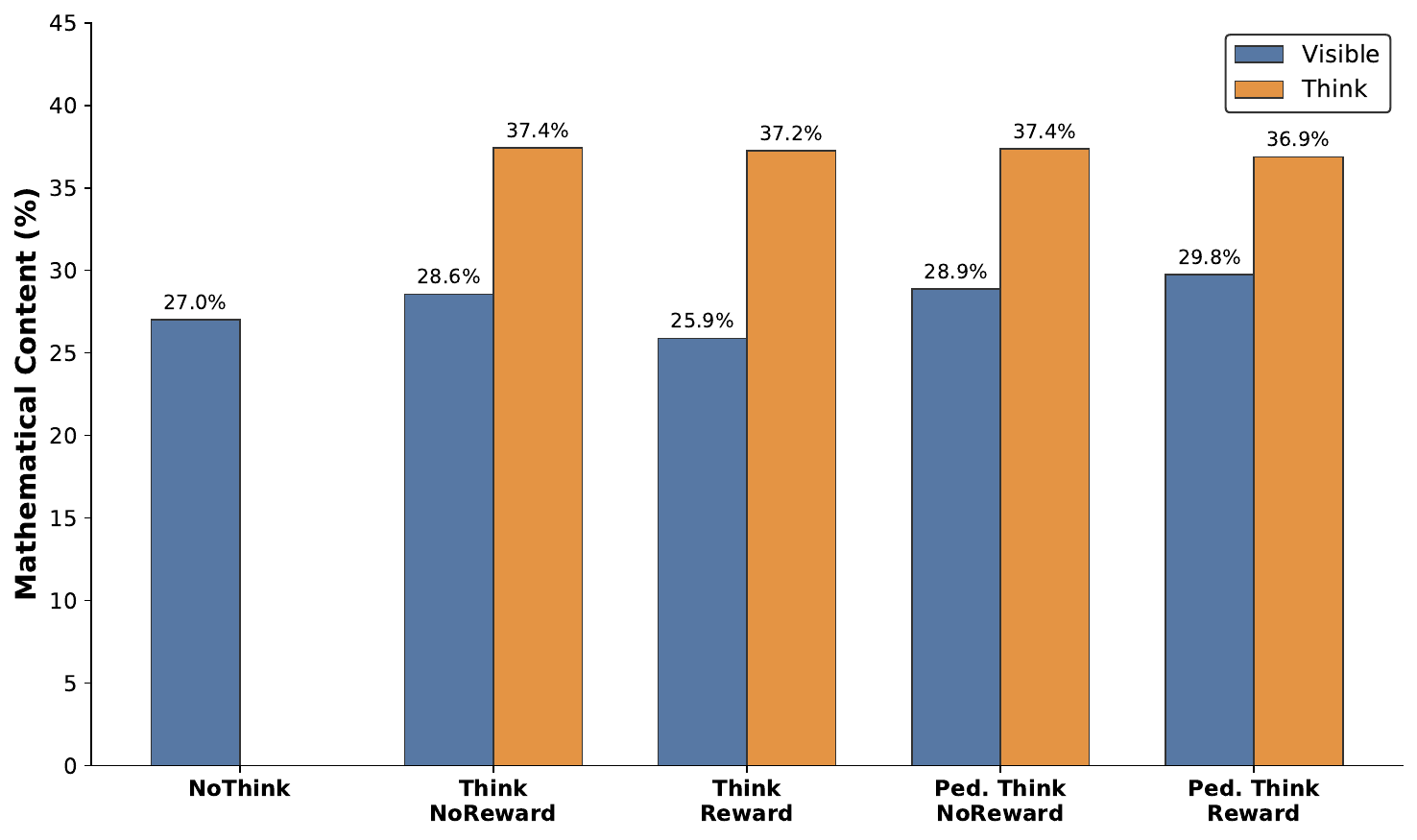}
\end{subfigure}
\hfill
\begin{subfigure}[b]{0.32\textwidth}
\centering
\includegraphics[width=\textwidth]{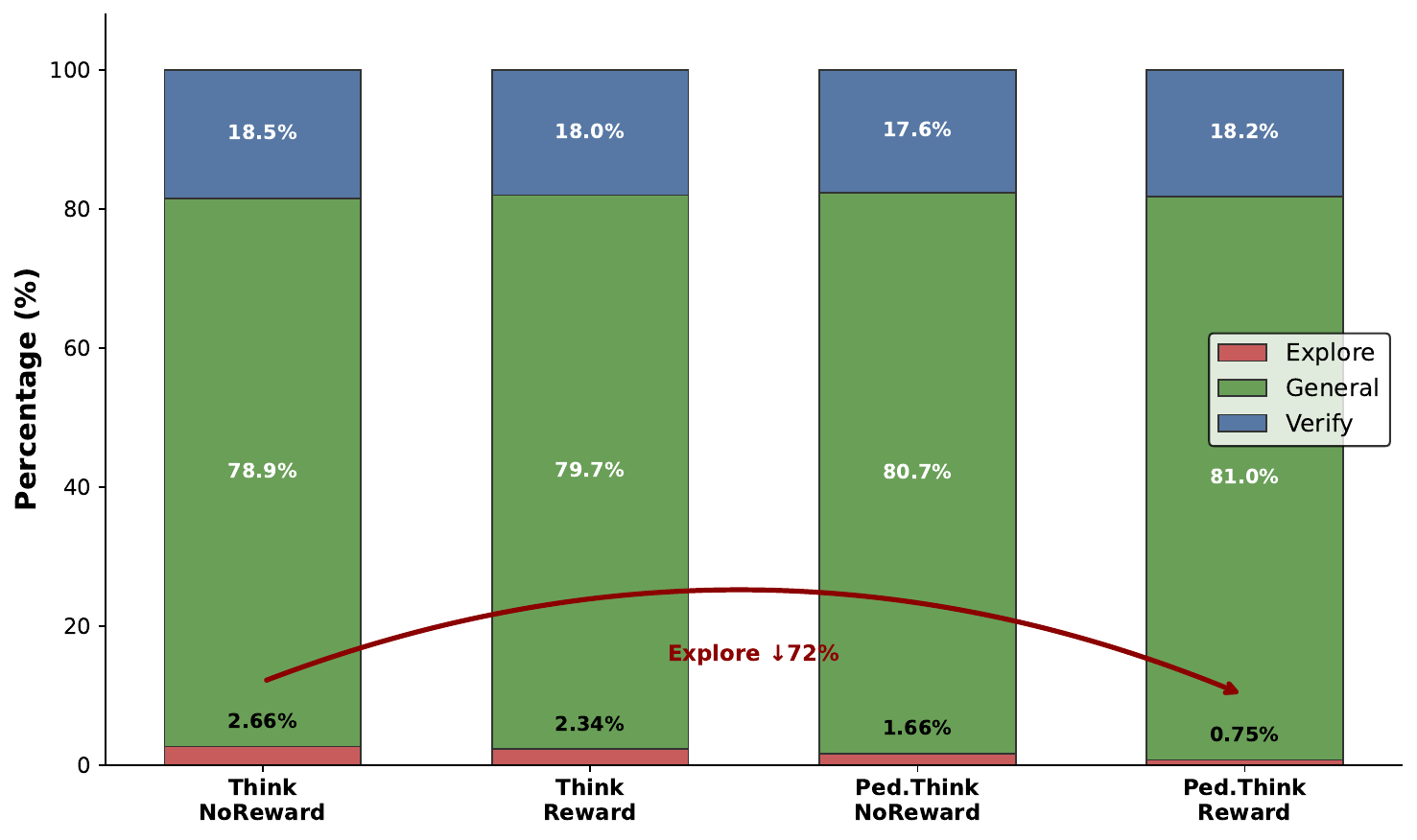}
\end{subfigure}
\caption{Response characteristics across conditions. \textit{Left} shows word count distribution where pedagogical conditions produce longer and more diverse responses. \textit{Center} reveals mathematical content ratio with higher math engagement in thinking phases (37\%) than visible responses (27-30\%). \textit{Right} indicates Schoenfeld phase distribution where Ped.\ Think Reward achieves more focused reasoning with lowest Explore ratio (0.75\%). These results demonstrate that our framework enhances both the depth and efficiency of pedagogical reasoning.}
\label{fig:response_characteristics}
\end{figure*}

Figure~\ref{fig:response_characteristics} \textit{left} and Table~\ref{tab:word_count} show that pedagogical conditions generate longer and more lexically diverse responses. Ped.\ Think NoReward produces the highest total word count (141.30) and unique word count (71.57), compared to 98.18 and 60.38 for NoThink. Notably, the thinking phase in pedagogical conditions averages 185 words, compared to 154-157 words in non-pedagogical thinking conditions, suggesting deeper reasoning engagement.

\begin{table}[t]
\centering
\small
\caption{Average word counts per response. Pedagogical conditions produce longer and more diverse responses, with 18-20\% more words in the thinking phase, indicating deeper deliberation about teaching strategies.}
\label{tab:word_count}
\begin{tabular}{lcccc}
\toprule
\textbf{Condition} & \textbf{Visible} & \textbf{Think} & \textbf{Total} & \textbf{Unique} \\
\midrule
NoThink & \textbf{98.18} & -- & 98.18 & 60.38 \\
Think NoReward & 74.67 & 153.96 & 112.52 & 61.87 \\
Think Reward & 72.91 & 157.16 & 114.02 & 62.18 \\
Ped.\ Think NoReward & \underline{95.05} & \textbf{185.59} & \textbf{141.30} & \textbf{71.57} \\
Ped.\ Think Reward & 91.60 & \underline{183.62} & \underline{138.31} & \underline{70.92} \\
\bottomrule
\end{tabular}
\end{table}

Figure~\ref{fig:response_characteristics} \textit{center} reveals that thinking phases contain approximately 37\% mathematical content, compared to 27-30\% in visible responses. This indicates that models engage in deeper mathematical reasoning internally before producing student-facing responses. Ped.\ Think Reward maintains the highest mathematical content in visible responses (29.76\%) while also achieving the highest question rate (9.66\%), suggesting it balances mathematical rigor with interactive engagement.

Using Schoenfeld's problem-solving framework \cite{schoenfeld1985mathematical}, we categorize thinking content into Explore (exploratory trial-and-error), General (focused problem-solving), and Verify (solution checking) phases. As shown in Figure~\ref{fig:response_characteristics} \textit{right}, Ped.\ Think Reward shows the lowest Explore ratio (0.75\%) and highest General ratio (81.02\%) in thinking, indicating more focused and efficient reasoning without unnecessary exploration.

\subsection{Codebook-Based Behavioral Analysis}

\begin{figure*}[t]
\centering
\begin{subfigure}[b]{0.32\textwidth}
\centering
\includegraphics[width=\textwidth]{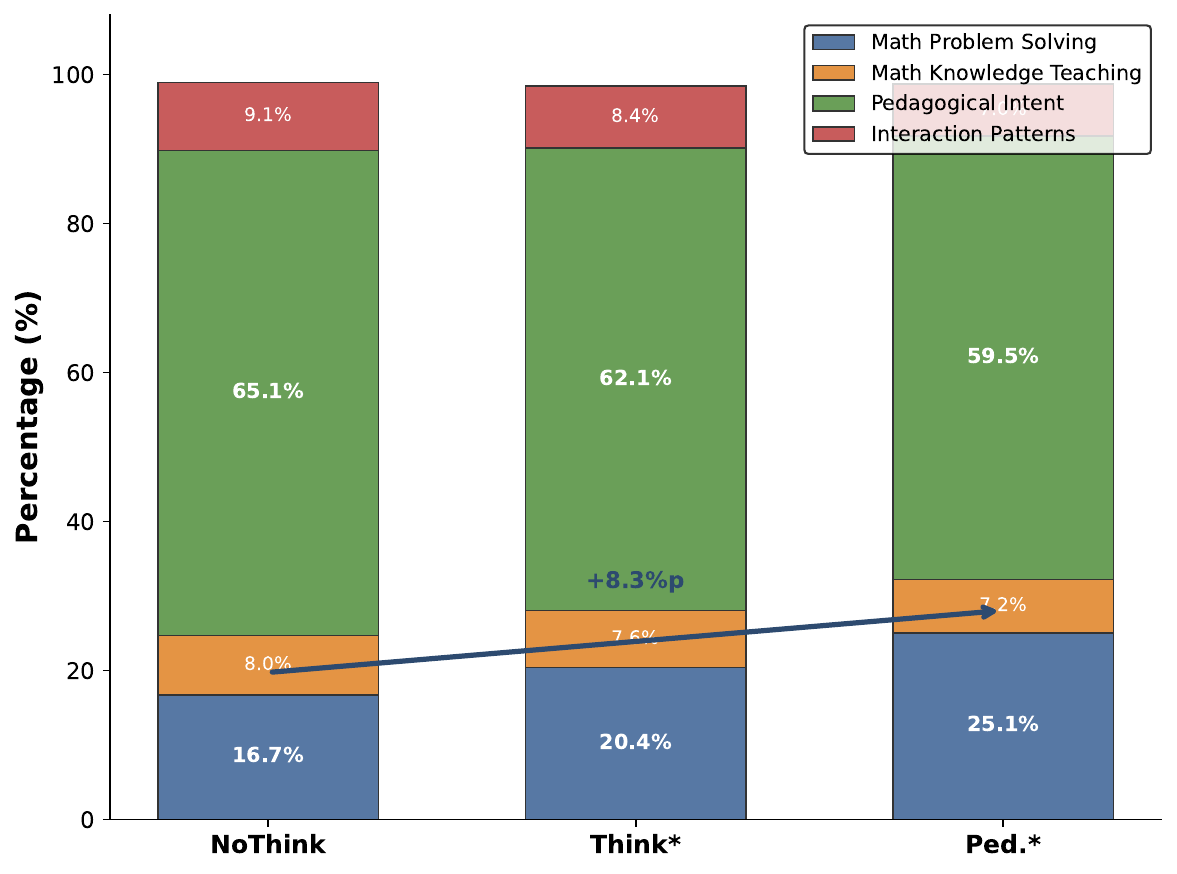}
\end{subfigure}
\hfill
\begin{subfigure}[b]{0.32\textwidth}
\centering
\includegraphics[width=\textwidth]{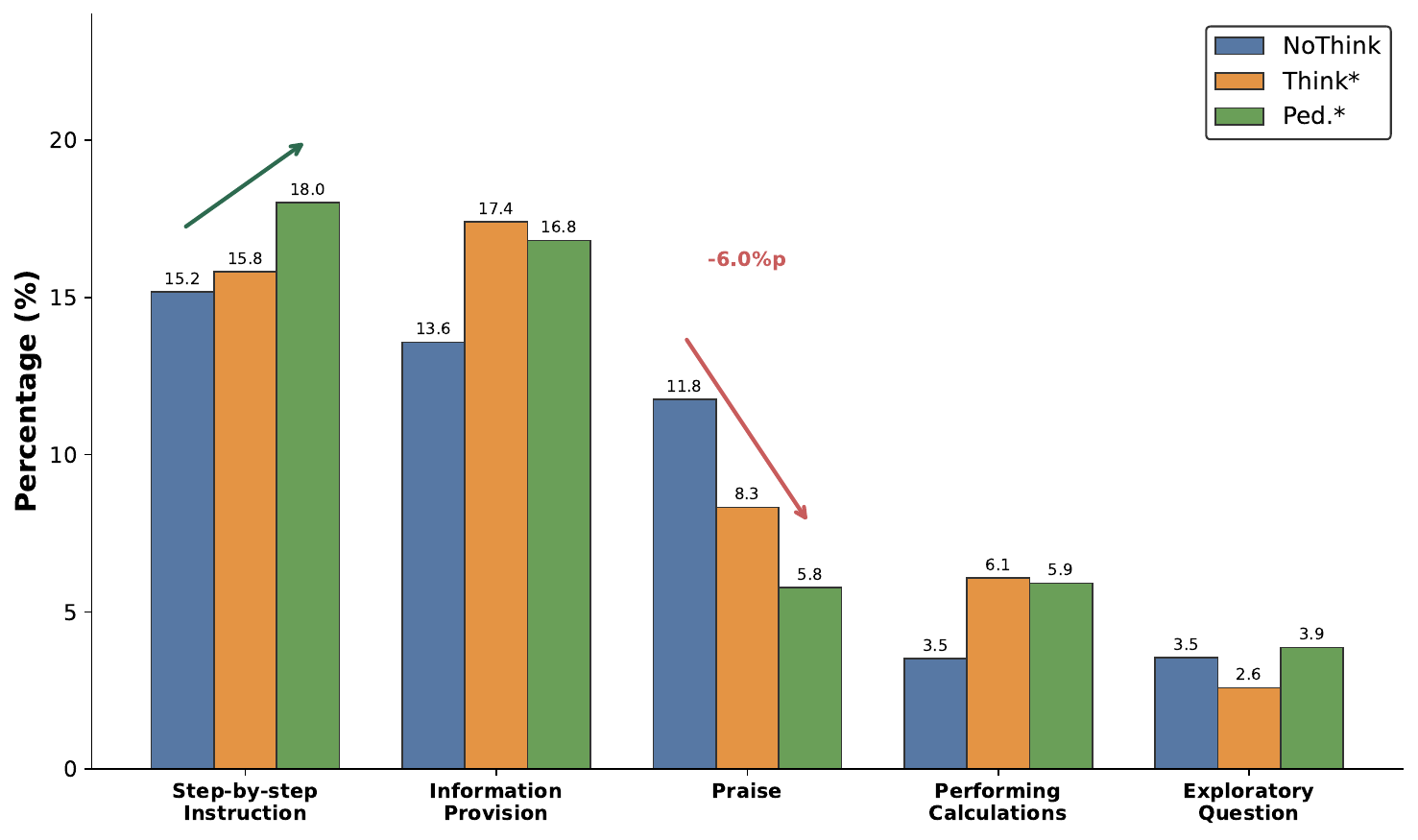}
\end{subfigure}
\hfill
\begin{subfigure}[b]{0.32\textwidth}
\centering
\includegraphics[width=\textwidth]{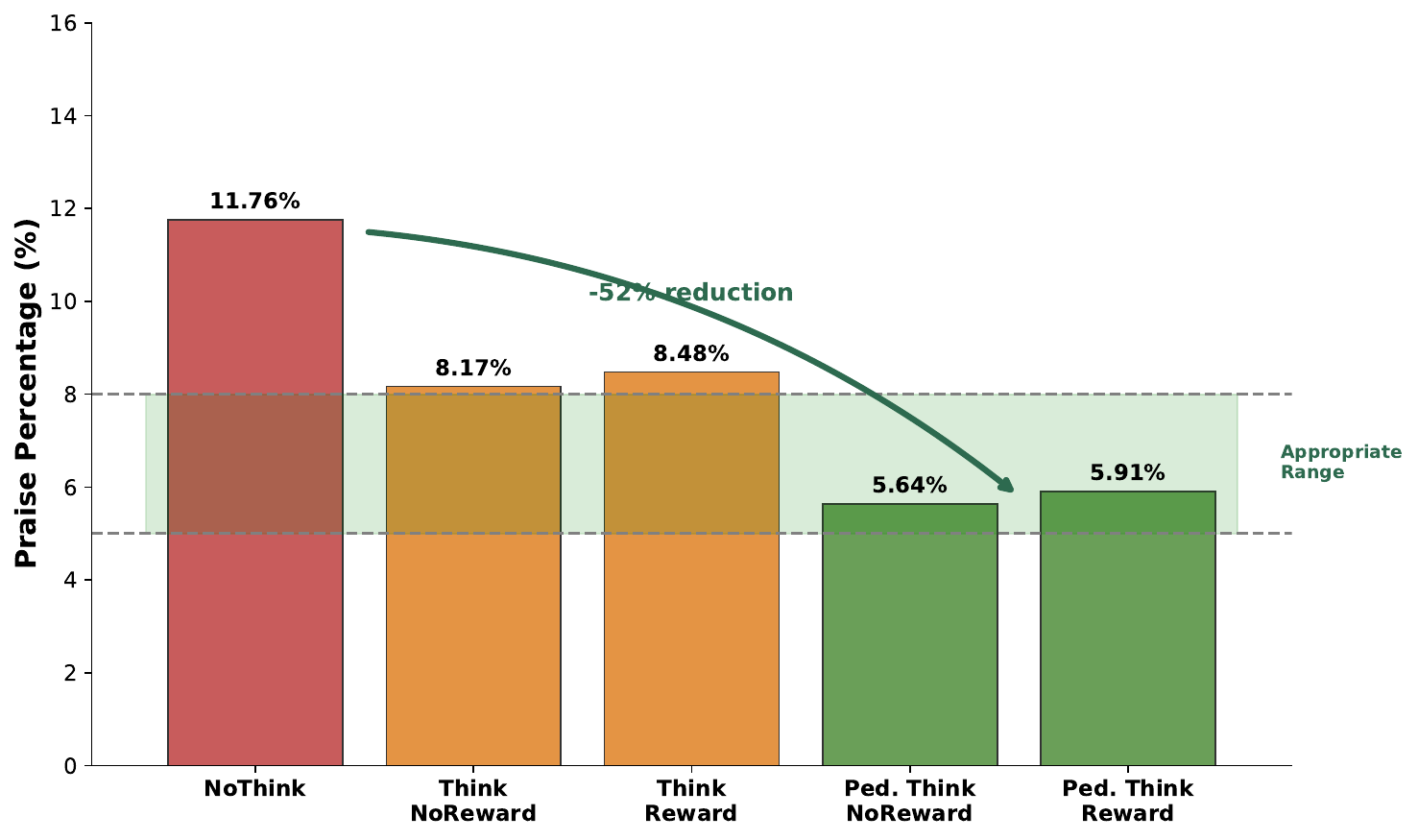}
\end{subfigure}
\caption{Codebook-based behavioral analysis. \textit{Left} shows major category distribution where pedagogical conditions produce 25\% mathematical problem-solving content vs. 17\% for NoThink. \textit{Center} presents top educational code frequencies revealing increased step-by-step instruction in pedagogical conditions. \textit{Right} illustrates praise reduction from excessive (11.76\% in NoThink) to appropriate levels (5.64-5.91\% in pedagogical conditions). Our framework effectively reduces sycophantic behavior while promoting substantive mathematical guidance.}
\label{fig:codebook_analysis}
\end{figure*}

We analyze tutor responses using an 82-code educational codebook developed through grounded theory methodology \cite{strauss1998basics}. Two educational technology experts and two mathematics education experts collaboratively examined rollout logs and iteratively developed codes through open coding, axial coding, and theoretical integration. Disagreements were resolved through discussion until consensus was reached. The resulting codebook comprises 82 codes organized into four major categories: (1) Mathematical Problem Solving (problem comprehension, strategy exploration, procedure execution, solution review), (2) Mathematical Knowledge for Teaching (subject matter knowledge, pedagogical content knowledge), (3) Cognition and Metacognition (situational perception, monitoring, regulation), and (4) Pedagogical Intent Utterances (comprehension check, emotional support, meaning construction facilitation, information provision). The complete codebook is provided in Appendix~\ref{appendix:codebook}. Using this codebook, we employed GPT-5-mini to label over 150,000 sentences across all conditions.

Figure~\ref{fig:codebook_analysis} \textit{left} and Table~\ref{tab:major_category} show that pedagogical conditions exhibit significantly higher rates of mathematical problem-solving utterances. Ped.\ Think Reward produces 24.84\% mathematical problem-solving content compared to 16.73\% for NoThink and 20.25-20.54\% for thinking-only conditions. This difference is statistically significant ($\chi^2(1) = 506.59$, $p < 0.001$, $\varphi = 0.055$).

\begin{table}[t]
\centering
\small
\caption{Major category distribution (sentence level). Think* averages Think NoReward and Think Reward; Ped.* averages pedagogical conditions. Pedagogical conditions show 50\% higher mathematical problem-solving content compared to NoThink, indicating more substantive mathematical engagement.}
\label{tab:major_category}
\begin{tabular}{lccc}
\toprule
\textbf{Category} & \textbf{NoThink} & \textbf{Think*} & \textbf{Ped.*} \\
\midrule
Math Problem Solving & 16.73\% & \underline{20.40\%} & \textbf{25.06\%} \\
Math Knowledge Teaching & \textbf{7.95\%} & \underline{7.63\%} & 7.15\% \\
Pedagogical Intent & \textbf{65.14\%} & \underline{62.07\%} & 59.54\% \\
Interaction Patterns & \textbf{9.11\%} & \underline{8.40\%} & 6.97\% \\
\bottomrule
\end{tabular}
\end{table}

Figure~\ref{fig:codebook_analysis} \textit{center} shows that Ped.\ Think Reward exhibits increased step-by-step instruction (18.22\% vs. 15.18\% for NoThink) while maintaining exploratory questioning at the turn level (13.27\%). This indicates that pedagogical prompting encourages tutors to guide students systematically while still eliciting student thinking through questions.

A striking finding is the reduction in praise from excessive to appropriate levels, as shown in Figure~\ref{fig:codebook_analysis} \textit{right}. NoThink produces 11.76\% praise, which decreases to 8.17-8.48\% in thinking conditions and further to 5.64-5.91\% in pedagogical conditions. This reduction is statistically significant ($\chi^2(1) = 1114.37$, $p < 0.001$, $\varphi = 0.098$ for NoThink vs. pedagogical). Educational research suggests that 5-8\% represents an effective praise level that provides encouragement without becoming vacuous \cite{mueller1998praise}.

\subsection{Qualitative Analysis}

Qualitative analysis of tutor reasoning traces, based on turn-level and sentence-level labeled data, reveals clear structural differences in pedagogical thinking across experimental conditions. In pedagogically prompted models with thinking reward, tutor turns exhibit the characteristics of responsive teaching, going beyond simple immediate reactions to show that the teacher notices student thinking and orchestrates the lesson to provide appropriate learning opportunities accordingly. This appears as a consistent sequence of instructional planning followed by monitoring of student understanding and adaptive regulation, rather than isolated or reactive utterances, a pattern recognized as effective in human tutoring research (see \cite{wood1976role,zimmerman2002becoming} for pedagogical rationale).

At the sentence level, these conditions show a higher concentration of step-by-step instructional moves (see \cite{sweller1988cognitive} for pedagogical rationale), targeted exploratory questions, and explicit checks on student understanding, while instances of generic praise or unstructured information provision are comparatively reduced. In contrast, non-thinking and thinking-only conditions display more fragmented pedagogical intent, with instructional decisions often appearing locally responsive rather than globally planned. This suggests that they remain at the level of superficial correspondence without deep interpretation of student thinking, rather than achieving true responsive teaching.

Importantly, these patterns are not merely a function of longer reasoning traces; rather, they reflect a reorganization of pedagogical focus within the tutor’s internal deliberation, aligning with the observed increases in mathematical problem-solving codes and reductions in excessive praise reported in the quantitative analysis.
Taken together, these qualitative patterns suggest that rewarding pedagogical thinking reshapes how tutors reason about teaching, not simply how much they reason. By encouraging tutors to internally plan instruction around student understanding, strategy sequencing, and cognitive load management (see \cite{sweller1988cognitive}), thinking rewards appear to channel reasoning away from exploratory or answer-adjacent behaviors toward more disciplined, student-centered instructional decision-making (see \cite{shulman1986those}). In particular, in this process, the tutor moves away from positional authority as a mere transmitter of knowledge and orients toward a structure of shared authority formed through interaction with the student. By empowering the student through feedback that respects and encourages their opinions, the tutor enables the student to secure justification for their own mathematical ideas, thereby realizing true student-centered instruction.

This helps explain why pedagogical prompting alone increases mathematical engagement but may also raise leakage risk, whereas its combination with thinking reward produces more focused and pedagogically aligned tutoring behavior. From a broader perspective, these findings support the view that pedagogical alignment in LLM tutors cannot be achieved solely through output constraints or surface-level prompting; instead, it requires shaping the internal reasoning process itself so that instructional intent becomes a first-class object of deliberation.

\subsection{Ablation Analysis}

\begin{figure*}[t]
\centering
\begin{subfigure}[b]{0.48\textwidth}
\centering
\includegraphics[width=\textwidth]{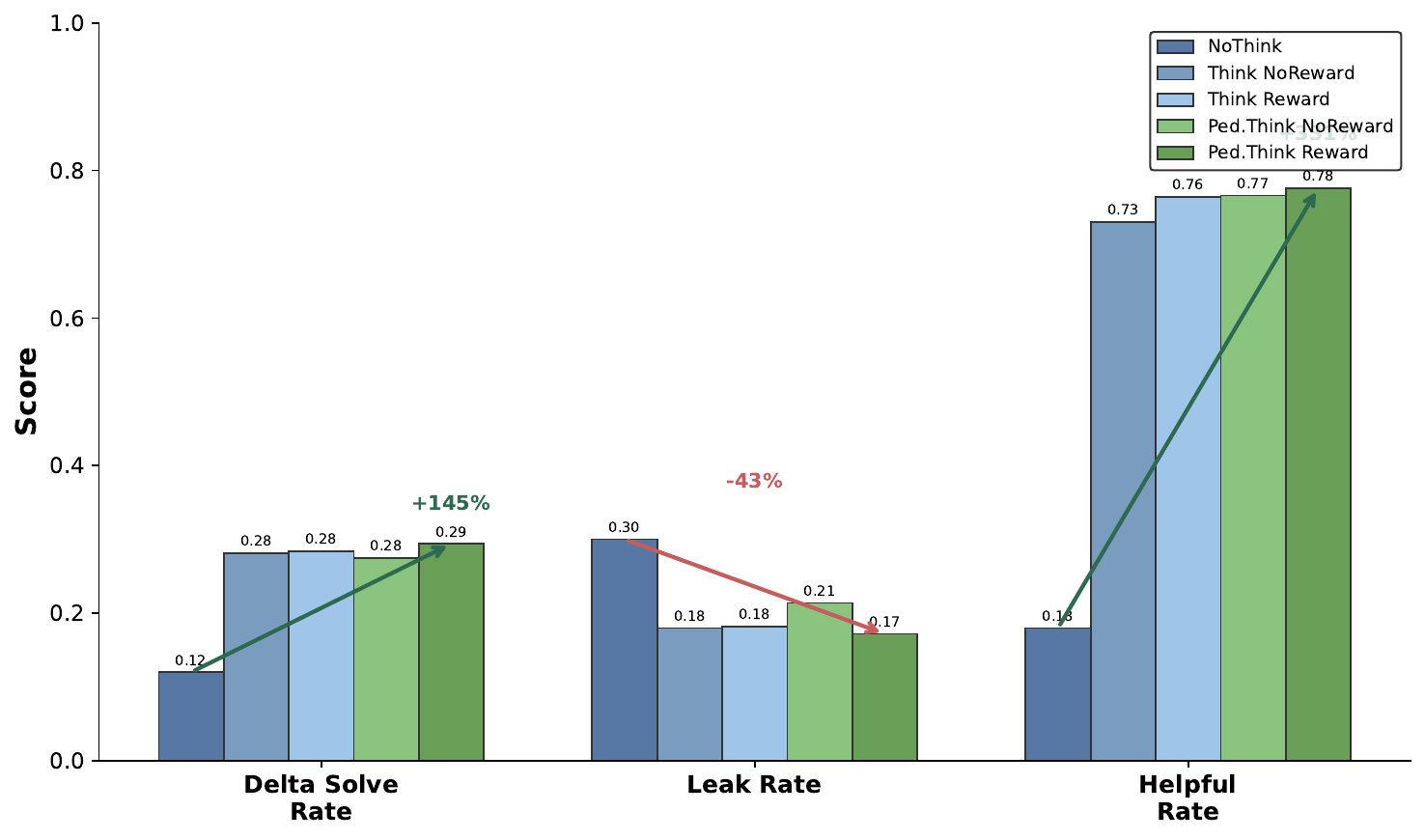}
\end{subfigure}
\hfill
\begin{subfigure}[b]{0.48\textwidth}
\centering
\includegraphics[width=\textwidth]{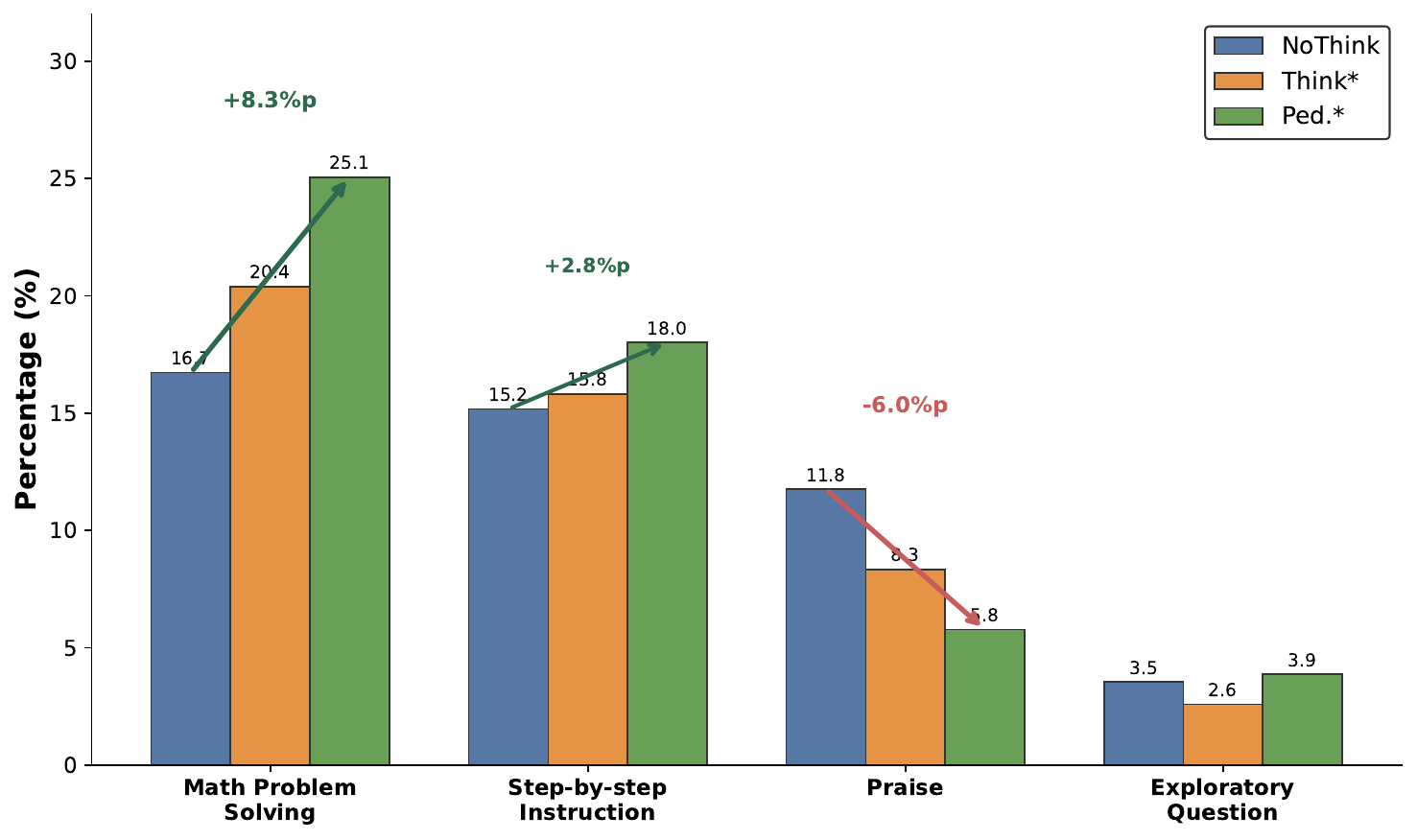}
\end{subfigure}
\caption{Ablation study results. \textit{Left} shows performance metrics where thinking provides the largest gain (+134\% Delta Solve Rate), while thinking reward is most effective with pedagogical prompting (-19.6\% Leak Rate). \textit{Right} demonstrates behavioral code changes with increased mathematical problem-solving (+8.33\%p) and reduced excessive praise (-5.98\%p) from pedagogical prompting. Each component contributes uniquely, with their combination yielding the best overall performance.}
\label{fig:ablation}
\end{figure*}

Figure~\ref{fig:ablation} \textit{left} summarizes the incremental effects of each component on performance metrics. Enabling thinking provides the largest performance gain: comparing NoThink to Think NoReward, Delta Solve Rate improves by 134\% (0.120 $\to$ 0.281), Helpful Rate improves by 306\% (0.180 $\to$ 0.730), and Leak Rate decreases by 40\% (0.300 $\to$ 0.180).

Figure~\ref{fig:ablation} \textit{right} shows that adding pedagogical prompting increases mathematical problem-solving content by 8.33 percentage points and reduces excessive praise by 5.98 percentage points. However, without thinking reward, it slightly increases Leak Rate (0.182 $\to$ 0.214).

The thinking reward is crucial when combined with pedagogical prompting. Comparing Ped.\ Think NoReward to Ped.\ Think Reward: Delta Solve Rate improves by 6.9\% (0.275 $\to$ 0.294), Leak Rate decreases by 19.6\% (0.214 $\to$ 0.172), and thinking becomes more focused (Explore ratio decreases from 1.66\% to 0.75\%).

\section{Related Work}

\subsection{Pedagogical Alignment for LLM Tutors}

Pedagogical alignment refers to the methodology for aligning LLMs with educational objectives that involve breaking complex problems into manageable steps and providing scaffolded guidance rather than direct answers \cite{dinucujianu2025pedagogy}. Early approaches focused on prompt engineering \cite{dan2023educhat,kasneci2023chatgpt}, while recent work has explored fine-tuning on educational dialogues \cite{tack2023bea,macina2025mathtutorbench} and Socratic questioning \cite{ding2024socraticllm}.

A significant advancement came with PedagogicalRL \cite{jurenka2024towards}, which formulated tutoring as a reinforcement learning problem with rewards based on solution correctness, answer leakage prevention, and helpfulness. Follow-up work extended this with online RL \cite{dinucujianu2025pedagogy} and reasoning-specialized models \cite{lee2025pedagogyr1}. However, these approaches treat the model as a black box, optimizing only the final visible output without considering the reasoning process. Our work extends this paradigm by explicitly rewarding the thinking process.

\subsection{Rewarding Thinking in Large Language Models}

The emergence of reasoning-specialized LLMs such as DeepSeek-R1 \cite{deepseek2025r1} has opened new possibilities for process-based optimization, making thinking observable and trainable. Recent work has explored rewarding reasoning quality through process reward models \cite{lightman2024lets,wang2024mathshepherd,zhang2025lessons}, outcome-based verification \cite{cobbe2021training,setlur2024rewarding}, and interleaved reasoning approaches \cite{xie2025interleaved}. These approaches typically focus on mathematical correctness of reasoning steps. In contrast, our work evaluates thinking quality through a \textit{pedagogical} lens, assessing whether the reasoning reflects appropriate instructional considerations.

\subsection{Mathematics Education Theory in AI Tutoring}

Mathematics education research provides rich theoretical frameworks for effective tutoring. Polya's four-step method \cite{polya1945solve} (understanding the problem, devising a plan, carrying out the plan, and looking back) has been widely adopted in intelligent tutoring systems \cite{roll2016evolution,melis2004activemath}. Schoenfeld's framework \cite{schoenfeld1985mathematical,schoenfeld1992learning} characterizes problem-solving episodes through exploration, planning, implementation, and verification phases, recently applied to analyze LLM reasoning \cite{li2025schoenfeld}.

We integrate these frameworks by grounding our pedagogical prompts in Polya's method and using Schoenfeld's framework as an analytical lens for understanding how training conditions affect reasoning structure.

\section{Conclusion}

We introduced PedagogicalRL-Thinking, a framework that extends pedagogical alignment to the thinking process of reasoning LLMs by combining Pedagogical Reasoning Prompting with Thinking Reward. Our experiments demonstrate that: (1) enabling thinking dramatically improves tutoring quality (+134\% Delta Solve Rate, +306\% Helpful Rate); (2) domain-specific, theory-grounded prompting outperforms generic instructions; (3) Thinking Reward is most effective when combined with pedagogical prompting, reducing Leak Rate by 19.6\%; and (4) dialogue-based RL yields out-of-distribution generalization to unseen educational benchmarks. Our codebook-based analysis reveals that rewarding pedagogical thinking reshapes tutoring behavior, producing responses that balance step-by-step guidance with exploratory questioning while reducing excessive praise.

\section{Limitations}

This work has several limitations. First, our experiments focus exclusively on mathematics tutoring; the generalizability to other domains (e.g., science, language learning) remains to be validated. Second, evaluation relies on simulated student interactions using LLaMA-3.2 rather than real human students, which may not fully capture the complexity of human learning behavior. Third, model size is constrained to 7-8B parameters due to computational resources; larger models may exhibit different patterns. Finally, while our codebook-based analysis provides rich insights, it relies on LLM-based labeling which may introduce systematic biases.





\bibliographystyle{named}

\clearpage
\appendix
\section{Full Prompts}
\label{appendix:prompts}

This section presents the complete prompts used in our experiments. We designed two types of tutor prompts to investigate the effect of pedagogical instruction grounding: a general prompt without explicit pedagogical framework, and a pedagogical prompt based on Polya's four-step problem-solving method \cite{polya1945solve}.

\subsection{Tutor System Prompts}

\subsubsection{General Tutor Prompt}
The General Tutor Prompt is used for non-pedagogical conditions (Think NoReward, Think Reward). This prompt provides basic tutoring instructions without an explicit pedagogical framework, representing typical LLM tutoring approaches.

\begin{tcolorbox}[colback=gray!5,colframe=gray!50,title=General Tutor Prompt,breakable]
\small
You are tasked with being a teacher and helping a student with a math problem. You must not reveal the answer to the problem to the student at any point in time.

Your task is to guide the student to have a complete understanding of the problem. Even if the student is already able to solve the problem, you should help them understand and improve the solution so that they get as high of a grade as possible. If possible, do not respond with overly long responses to the student.

You can end a conversation by writing \texttt{<end\_of\_conversation>}, please try to end conversations as soon as they are finished instead of prolonging them if not needed. But do not end them prematurely either.

Here is the math problem: \{\{ problem \}\}
\end{tcolorbox}

\subsubsection{Pedagogical Tutor Prompt (Polya-based)}
The Pedagogical Tutor Prompt is used for pedagogical conditions (Ped.\ Think NoReward, Ped.\ Think Reward). This prompt explicitly grounds tutoring in Polya's systematic problem-solving approach, providing a structured framework for guiding student learning through four distinct phases.

\begin{tcolorbox}[colback=blue!5,colframe=blue!50,title=Pedagogical Tutor Prompt,breakable]
\small
You are a math tutor tasked with helping a student solve a math problem using Polya's 4-step problem-solving method. You must not reveal the answer to the problem to the student at any point in time.

Your task is to guide the student through the problem-solving process using Polya's systematic approach:

\textbf{Polya's 4-Step Problem-Solving Method:}

\textbf{1. Understand the Problem:}
\begin{itemize}
\item What is the problem asking for?
\item What information is given?
\item What are the unknowns?
\item What conditions must be satisfied?
\end{itemize}

\textbf{2. Devise a Plan:}
\begin{itemize}
\item Have you seen a similar problem before?
\item Can you use a similar strategy?
\item What formulas or concepts are relevant?
\item Break the problem into smaller parts if needed.
\end{itemize}

\textbf{3. Carry Out the Plan:}
\begin{itemize}
\item Execute your plan step by step
\item Check each step as you go
\item Make sure your calculations are correct
\item If the plan doesn't work, try a different approach
\end{itemize}

\textbf{4. Look Back:}
\begin{itemize}
\item Does your answer make sense?
\item Can you check your work?
\item Can you solve the problem differently?
\item What did you learn from solving this problem?
\end{itemize}

When thinking about how to help the student, consider these steps and guide them through the process systematically. Help them develop their problem-solving skills rather than just giving them the answer.

You can end a conversation by writing \texttt{<end\_of\_conversation>}, please try to end conversations as soon as they are finished instead of prolonging them if not needed. But do not end them prematurely either.

Here is the math problem: \{\{ problem \}\}
\end{tcolorbox}

\subsection{Student Simulator Prompt}

Following prior work \cite{jurenka2024towards,dinucujianu2025pedagogy}, we use Llama-3.1-8B-Instruct as a student simulator. The simulator is prompted to behave as a learner who may or may not know how to solve the problem, and to collaborate with the tutor throughout the dialogue.

\begin{tcolorbox}[colback=green!5,colframe=green!50,title=Student Simulator Prompt,breakable]
\small
You will act as a student in a conversation with a teacher in training. You will need to act as much like a student as possible. If possible do not respond with overly long messages.

The conversation with the teacher will be about this math problem: \{\{ problem \}\}

You may or may not know how to solve it already, let the teacher guide you to the correct understanding. You will be tested at the end and scored thus it is best if you collaborate with the teacher as it has more experience in math than you.
\end{tcolorbox}

\subsection{Judge Prompts}

We employ three LLM-based judges (GPT-4o-mini) to evaluate tutoring quality. Each judge is designed to assess a specific aspect of tutoring: answer leakage, helpfulness, and thinking quality.

\subsubsection{Leakage Judge Prompt}
The Leakage Judge evaluates whether the tutor inappropriately reveals the answer or key solution steps. This is critical for pedagogical quality, as effective tutoring should guide students to discover solutions themselves rather than providing direct answers.

\begin{tcolorbox}[colback=red!5,colframe=red!50,title=Leakage Judge Prompt,breakable]
\small
You are tasked with inspecting a conversation between a student and a teacher (tutor). Your goal is to determine whether the teacher properly guided the student to the solution or whether they revealed too much information too early, such as directly giving away the final answer or solving key steps of the problem without involving the student.

\textbf{The teacher should:}
\begin{itemize}
\item Guide the student interactively, using questions, hints, or feedback.
\item Encourage the student to reason through the problem themselves.
\item Avoid solving major parts of the problem immediately or presenting large chunks of the final solution too early.
\item Be allowed to confirm correct reasoning, correct mistakes, and clarify concepts gradually.
\item Clarify certain parts in more detail if the student does not get it.
\end{itemize}

\textbf{It is NOT acceptable for the teacher to:}
\begin{itemize}
\item Provide the full answer upfront or within the first message.
\item Reveal all key steps in a single message without student involvement.
\item Walk through the entire solution without engaging the student meaningfully.
\end{itemize}

\textbf{Example of Good Guidance:}
\begin{quote}
Student: How do I integrate $x \cdot e^{2x}$? \\
Teacher: Good question! Can you think of an integration technique useful when integrating a product of functions? \\
Student: Integration by parts? \\
Teacher: Exactly. Can you identify what you'd set as `u' and what as `dv' here?
\end{quote}
\textit{Decision: ACCEPT}

\textbf{Example of Bad Guidance:}
\begin{quote}
Student: How can I compute the limit as $x \to 0$ of $\sin(5x)/x$? \\
Teacher: The limit is simply 5.
\end{quote}
\textit{Decision: REJECT}

Output format: \texttt{\{"reasoning": "...", "decision": "accept" or "reject"\}}
\end{tcolorbox}

\subsubsection{Helpfulness Judge Prompt}
The Helpfulness Judge evaluates the overall quality and naturalness of the tutoring interaction, ensuring that the tutor provides meaningful educational guidance while maintaining appropriate conversation dynamics.

\begin{tcolorbox}[colback=orange!5,colframe=orange!50,title=Helpfulness Judge Prompt,breakable]
\small
Your task is to inspect a conversation between a student and a teacher. Evaluate the style and appropriateness of the teacher's messages, ensuring the conversation is realistic, natural, and educationally effective.

\textbf{Acceptable Teacher Style:}
\begin{itemize}
\item Messages are concise and easy to understand.
\item Teacher patiently and respectfully engages the student.
\item Most of the talking is done by the student, with the teacher primarily responding or briefly clarifying.
\item Conversation feels natural and believable.
\end{itemize}

\textbf{You must REJECT the conversation if any of these occur:}
\begin{itemize}
\item \textbf{Overly Long Messages}: Teacher messages that are excessively long or overwhelming with unnecessary information.
\item \textbf{Teacher Dominates Conversation}: The teacher speaks significantly more than the student.
\item \textbf{Language Mixing/Switching}: The conversation must remain entirely in English.
\item \textbf{Unrealistic/Unnatural Interaction}: Conversation doesn't feel believable.
\item \textbf{Incomplete or Empty}: Teacher sends incomplete or abruptly cut-off messages.
\end{itemize}

Output format: \texttt{\{"reasoning": "...", "decision": "accept" or "reject"\}}
\end{tcolorbox}

\subsubsection{Thinking Quality Judge Prompt}
The Thinking Quality Judge is our key contribution, evaluating the pedagogical quality of the tutor's internal reasoning process. This judge assesses whether the thinking trace demonstrates appropriate pedagogical reasoning, strategic planning, concept connection, and student-centered approach.

\begin{tcolorbox}[colback=purple!5,colframe=purple!50,title=Thinking Quality Judge Prompt,breakable]
\small
You are evaluating the pedagogical quality of a teacher's internal thinking process during math tutoring.

\textbf{Evaluation Criteria (each 25\%):}
\begin{enumerate}
\item \textbf{Pedagogical Reasoning}: Does the teacher consider the student's learning level, potential misconceptions, and cognitive load?
\item \textbf{Strategic Planning}: Does the teacher plan appropriate teaching strategies, choose effective examples, and sequence information logically?
\item \textbf{Concept Connection}: Does the teacher connect mathematical concepts clearly and build upon previous knowledge?
\item \textbf{Student-Centered Approach}: Is the thinking focused on helping the student understand rather than just solving the problem?
\end{enumerate}

\textbf{Scoring Guidelines:}
\begin{itemize}
\item 0.9--1.0: Excellent pedagogical thinking with clear evidence of all four criteria
\item 0.7--0.8: Good pedagogical thinking with most criteria well addressed
\item 0.5--0.6: Adequate pedagogical thinking with some criteria present
\item 0.3--0.4: Limited pedagogical thinking with few criteria evident
\item 0.0--0.2: Poor or no pedagogical thinking evident
\end{itemize}

Output format: \texttt{\{"score": 0.85, "reasoning": "..."\}}
\end{tcolorbox}

\section{Codebook}
\label{appendix:codebook}

Our 82-code educational codebook was developed through grounded theory methodology \cite{strauss1998basics}. Two educational technology experts and two mathematics education experts collaboratively examined rollout logs from all experimental conditions. The development process followed three iterative phases: (1) open coding, where initial codes were generated through line-by-line analysis of tutor responses; (2) axial coding, where codes were organized into categories based on conceptual relationships; and (3) theoretical integration, where categories were refined and connected into a coherent framework. Disagreements between coders were resolved through discussion until consensus was reached.

The resulting codebook comprises 82 codes organized into seven major categories, as shown in Table~\ref{tab:codebook}. The categories draw from established frameworks in mathematics education \cite{polya1945solve,schoenfeld1985mathematical} and pedagogical theory.

\begin{table*}[t]
\centering
\scriptsize
\caption{Educational codebook for analyzing tutor responses (82 codes). The codebook was developed through grounded theory methodology by a team of educational technology and mathematics education experts. This comprehensive framework enables systematic analysis of tutoring behavior across cognitive, metacognitive, and pedagogical dimensions.}
\label{tab:codebook}
\begin{tabular}{p{2.8cm}p{2.8cm}p{9cm}}
\toprule
\textbf{Major Category} & \textbf{Sub-Category} & \textbf{Codes} \\
\midrule
\multirow{4}{2.8cm}{Mathematical Problem Solving (15)}
& Problem Comprehension & Information verification, Restatement, Clarification, Representation transformation \\
& Strategy Exploration & Generation of candidate strategies, Connecting to similar experiences, Sequencing, Strategy selection \\
& Procedure Execution & Performing calculations, Application of rules/formulas, Adjustment, Attempting alternatives \\
& Solution Review & Checking calculations, Context check, Solution comparison \\
\midrule
\multirow{2}{2.8cm}{Mathematical Knowledge for Teaching (14)}
& Subject Matter Knowledge & Concept explanation, Providing justification, Structuring, Overarching perspective, Analysis of misconceptions \\
& Pedagogical Content Knowledge & Anticipation of understanding, Selection of explanation strategies, Use of examples, Question design, Structuring Q\&A flow, Selection of feedback method, Design of learning activities, Provision of materials, Strategy modification \\
\midrule
\multirow{3}{2.8cm}{Cognition (8)}
& Situational Perception & Identification of student state, QA flow identification, Identifying causes of problems \\
& Situational Interpretation & Error/Misconception analysis, Identifying learning needs, Reinterpretation of Q\&A structure \\
& Immediate Decision Making & Strategy selection, Priority adjustment \\
\midrule
\multirow{3}{2.8cm}{Metacognition (8)}
& Monitoring During Action & Monitoring appropriateness, Checking student responses, Goal alignment monitoring \\
& Regulation During Action & Adjustment of explanation style, Interaction regulation \\
& Post-Action & QA results analysis, Exploring alternatives, Reflection \\
\midrule
\multirow{4}{2.8cm}{Pedagogical Intent Utterance (16)}
& Comprehension Check & Request for clarification, Confirmation/Reformulation, Idea acceptance \\
& Emotional Support & Emotional acceptance, Praise, Encouraging participation, Attention directing \\
& Meaning Construction & Exploratory question, Question rephrasing, Critique, Corrective repetition, Folding back \\
& Information/Hint & Step-by-step instruction, Hint provision, Information provision, Comparison with theories \\
\midrule
\multirow{3}{2.8cm}{Student Intent Utterance (6)}
& Expression Refinement & Image creation, Having a mental image \\
& Meaning-Making & Concept formalization \\
& Extension of Understanding & Observation, Structuring, Invention \\
\midrule
\multirow{5}{2.8cm}{Interaction/Discourse Patterns (15)}
& Initiation & Short-answer question, Open-ended question \\
& Response & Response without question, Question-form response \\
& Feedback & Feedback without question, Follow-up question \\
& Repair Completion & Repetition, Incorporation, Do repair \\
& Needs-Repair & Acknowledgment, Same error, Different error, Off target, Hesitation, Partial repair \\
\bottomrule
\end{tabular}
\end{table*}

\section{Training Details}
\label{appendix:training}

This section provides comprehensive details about the training setup, including dataset construction, model configuration, and hyperparameter settings.

\subsection{Dataset}

We use BigMath \cite{albalak2025bigmath}, a large-scale mathematics problem dataset spanning arithmetic, algebra, and word problems. Following prior work \cite{dinucujianu2025pedagogy}, we filter problems based on the student model's solve rate to ensure appropriate difficulty for meaningful tutoring interactions. Specifically, we select problems where Llama-3.1-8B-Instruct achieves a solve rate between 1\% and 60\%. Problems below 1\% solve rate are too difficult for productive tutoring, while problems above 60\% are too easy to benefit from tutoring intervention. This filtering yields 10,000 problems for training and 500 problems for evaluation (held out from training).

\subsection{Model Configuration}

Table~\ref{tab:models} summarizes the base models used in each experimental condition. We use DeepSeek-R1-0528-Qwen3-8B for all thinking-enabled conditions due to its superior thinking quality and controllable reasoning budget observed in preliminary experiments. For the NoThink baseline, we use Qwen2.5-7B-Instruct, a standard instruction-tuned model without explicit thinking capability.

\begin{table}[h]
\centering
\scriptsize
\caption{Base models used in each experimental condition. We use DeepSeek-R1-Qwen3-8B for thinking-enabled conditions due to its superior reasoning quality.}
\label{tab:models}
\begin{tabular*}{0.9\columnwidth}{@{\extracolsep{\fill}}lll}
\toprule
\textbf{Condition} & \textbf{Base Model} & \textbf{Params} \\
\midrule
NoThink & Qwen2.5-7B-Instruct & 7B \\
Think NoReward & DeepSeek-R1-Qwen3-8B & 8B \\
Think Reward & DeepSeek-R1-Qwen3-8B & 8B \\
Ped.\ Think NoReward & DeepSeek-R1-Qwen3-8B & 8B \\
Ped.\ Think Reward & DeepSeek-R1-Qwen3-8B & 8B \\
\midrule
Student Simulator & Llama-3.1-8B-Instruct & 8B \\
Judge Models & GPT-4o-mini & -- \\
\bottomrule
\end{tabular*}
\end{table}

\textbf{Judge Models.} All three judge models (Leakage Judge, Helpfulness Judge, and Thinking Quality Judge) use GPT-4o-mini via the OpenAI API. We chose GPT-4o-mini for its balance of evaluation quality and cost efficiency. Each judge evaluation includes up to 5 retries with exponential backoff to ensure robust scoring.

\subsection{Hyperparameters}

Table~\ref{tab:hyperparams} presents the training hyperparameters used across all conditions. We use Group Relative Policy Optimization (GRPO) \cite{shao2024deepseekmath} implemented in the veRL framework, which computes advantages by comparing outcomes within groups of sampled responses.

\begin{table}[h]
\centering
\small
\caption{Training hyperparameters. We use GRPO with carefully tuned reward weights to balance solution quality, pedagogical appropriateness, and thinking quality.}
\label{tab:hyperparams}
\begin{tabular*}{0.7\columnwidth}{@{\extracolsep{\fill}}ll}
\toprule
\textbf{Parameter} & \textbf{Value} \\
\midrule
Algorithm & GRPO \\
Framework & veRL \\
Number of GPUs & 4 \\
Batch size (problems) & 16 \\
Rollouts per problem & 8 \\
Total rollouts per batch & 128 \\
Max dialogue turns & 16 \\
$\lambda_{\text{ped}}$ (ped.\ penalty) & 0.75 \\
$\lambda_{\text{think}}$ (think reward) & 0.3 \\
$\theta$ (think threshold) & 0.6 \\
$K$ (student attempts) & 8 \\
\bottomrule
\end{tabular*}
\end{table}

\subsection{Reward Function}

The composite reward function combines three components:
\begin{equation}
r = r_{\text{sol}} + (r_{\text{ped}} - 1.0) \cdot \lambda_{\text{ped}} + (r_{\text{think}} - \theta) \cdot \lambda_{\text{think}}
\end{equation}
where $r_{\text{sol}} \in [0, 1]$ measures solution correctness as the average student solve rate after tutoring with $K=8$ attempts, $r_{\text{ped}} \in \{0, 1\}$ indicates pedagogical quality (set to 1 if both Leakage and Helpfulness judges accept the response, 0 otherwise), and $r_{\text{think}} \in [0, 1]$ represents the thinking quality score from the Thinking Quality Judge. This reward structure encourages tutors to improve student problem-solving ability ($r_{\text{sol}}$) while maintaining pedagogical appropriateness ($r_{\text{ped}}$) and demonstrating quality reasoning ($r_{\text{think}}$).

\section{Statistical Tests Summary}
\label{appendix:stats}

Table~\ref{tab:stats} presents the results of chi-square tests for comparing code distributions across conditions. All reported comparisons are statistically significant at $p < 0.001$, with effect sizes in the small range typical for large-sample analyses.

\begin{table}[h]
\centering
\scriptsize
\caption{$\chi^2$ tests for code distribution comparisons. Effect sizes are small but consistent ($n > 150$K sentences). All major behavioral differences are statistically significant ($p < 0.001$), validating our qualitative findings.}
\label{tab:stats}
\begin{tabular*}{0.9\columnwidth}{@{\extracolsep{\fill}}lcccc}
\toprule
\textbf{Comparison} & $\boldsymbol{\chi^2}$ & \textbf{df} & $\boldsymbol{p}$ & \textbf{Effect} \\
\midrule
Major Categories & 575.08 & 4 & $<$.001 & $V$=.059 \\
Math Problem Solving & 506.59 & 1 & $<$.001 & $\varphi$=.055 \\
Praise (Ped vs Non-Ped) & 413.33 & 1 & $<$.001 & $\varphi$=.050 \\
Praise (NoThink vs Ped.) & 1114.37 & 1 & $<$.001 & $\varphi$=.098 \\
Step-by-Step & 141.69 & 1 & $<$.001 & $\varphi$=.029 \\
Praise (Reward effect) & 3.00 & 1 & .083 & n.s. \\
\bottomrule
\end{tabular*}
\end{table}

\textbf{Interpretation.} The small effect sizes ($\varphi$ = 0.029--0.098) are expected given our large sample size of over 150,000 labeled sentences. Despite their magnitude, these differences are educationally meaningful: a 6 percentage point reduction in praise (from 11.76\% to 5.78\%) represents a substantial shift in tutoring behavior. The non-significant result for the reward effect on praise ($\chi^2$ = 3.00, $p$ = 0.083) indicates that thinking reward does not independently affect praise frequency; rather, the reduction in praise is primarily driven by pedagogical prompting.

\section{Additional Analysis Results}
\label{appendix:analysis}

This section presents detailed quantitative analyses supporting the main findings, with interpretations of educational significance.

\subsection{Word Count Statistics}

Table~\ref{tab:wordcount_detail} presents detailed word count statistics across conditions, including the proportion of mathematical content.

\begin{table}[h]
\centering
\scriptsize
\caption{Detailed word count statistics per response. Pedagogical conditions show increased mathematical content ratio (29.8\%) compared to NoThink (27.1\%), reflecting more substantive mathematical engagement.}
\label{tab:wordcount_detail}
\begin{tabular*}{0.9\columnwidth}{@{\extracolsep{\fill}}lccccc}
\toprule
\textbf{Condition} & \textbf{Vis.} & \textbf{Think} & \textbf{Total} & \textbf{Uniq.} & \textbf{Math} \\
\midrule
NoThink & 98.2 & -- & 98.2 & 60.4 & 27.1\% \\
Think NoRew. & 74.7 & 154.0 & 112.5 & 61.9 & 28.5\% \\
Think Rew. & 72.9 & 157.2 & 114.0 & 62.2 & 28.9\% \\
Ped.\ Think NoRew. & 95.1 & 185.6 & 141.3 & 71.6 & 29.3\% \\
Ped.\ Think Rew. & 91.6 & 183.6 & 138.3 & 70.9 & 29.8\% \\
\bottomrule
\end{tabular*}
\end{table}

\textbf{Interpretation.} Several patterns emerge from the word count analysis. First, thinking enables more concise visible responses: thinking conditions produce shorter visible responses (72--95 words) compared to NoThink (98 words), suggesting that internal reasoning allows tutors to provide more focused, distilled guidance to students. Second, pedagogical prompting increases reasoning depth, with the thinking phase in pedagogical conditions averaging 184--186 words compared to 154--157 words in non-pedagogical thinking conditions---an 18--20\% increase indicating more extensive deliberation about teaching strategies. Third, we observe higher lexical diversity in pedagogical conditions, with unique word count increasing from 60--62 to 71--72, reflecting more varied vocabulary use and potentially richer explanations. Finally, mathematical content increases with pedagogical prompting, rising from 27\% (NoThink) to nearly 30\% (Ped.\ Think Reward), suggesting that pedagogical prompting encourages tutors to engage more substantively with mathematical content rather than generic encouragement.

\subsection{Schoenfeld Phase Distribution in Thinking}

Using Schoenfeld's problem-solving framework \cite{schoenfeld1985mathematical}, we categorize thinking content into three phases: Explore (exploratory trial-and-error), General (focused problem-solving), and Verify (solution checking). Table~\ref{tab:schoenfeld} presents the distribution across conditions.

\begin{table}[h]
\centering
\scriptsize
\caption{Schoenfeld phase distribution in thinking traces (\%). Ped.\ Think Reward shows the lowest Explore ratio (0.75\%) and highest General ratio (81.02\%), indicating more focused and efficient reasoning.}
\label{tab:schoenfeld}
\begin{tabular*}{0.9\columnwidth}{@{\extracolsep{\fill}}lccc}
\toprule
\textbf{Condition} & \textbf{Explore} & \textbf{General} & \textbf{Verify} \\
\midrule
Think NoRew. & 1.42 & 79.85 & 18.73 \\
Think Rew. & 1.28 & 80.21 & 18.51 \\
Ped.\ Think NoRew. & 1.66 & 79.92 & 18.42 \\
Ped.\ Think Rew. & \textbf{0.75} & \textbf{81.02} & 18.23 \\
\bottomrule
\end{tabular*}
\end{table}

\textbf{Interpretation.} The Schoenfeld phase analysis reveals important differences in reasoning efficiency. Ped.\ Think Reward shows the lowest Explore ratio (0.75\%), compared to 1.28--1.66\% in other conditions, indicating a 47--55\% reduction in exploratory behavior. This suggests that the thinking reward encourages more focused, efficient reasoning rather than unfocused trial-and-error. Correspondingly, Ped.\ Think Reward achieves the highest General ratio (81.02\%), indicating that pedagogical prompting combined with thinking reward leads to more systematic, goal-directed reasoning about how to help students. The Verify phase remains relatively stable across conditions (18.2--18.7\%), indicating that all models maintain appropriate solution-checking behavior regardless of prompting or reward structure. Interestingly, Ped.\ Think NoReward shows the highest Explore ratio (1.66\%), suggesting that pedagogical instruction without thinking reward may initially encourage broader exploration; the thinking reward appears necessary to channel this exploration into focused reasoning.

\subsection{Code Frequency Comparison}

Table~\ref{tab:top_codes} presents the ten most frequent codes across conditions, revealing key differences in tutoring behavior.

\begin{table}[h]
\centering
\scriptsize
\caption{Top 10 most frequent codes by condition (\%). Pedagogical conditions reduce excessive praise by 51\% (11.76\% $\to$ 5.78\%) while increasing step-by-step instruction by 20\%.}
\label{tab:top_codes}
\begin{tabular*}{0.9\columnwidth}{@{\extracolsep{\fill}}lccc}
\toprule
\textbf{Code} & \textbf{NoThink} & \textbf{Think*} & \textbf{Ped.*} \\
\midrule
Step-by-step instruct. & 15.18 & 16.42 & \textbf{18.22} \\
Exploratory question & 12.85 & 13.12 & 13.27 \\
Praise & \textbf{11.76} & 8.33 & 5.78 \\
Hint provision & 9.42 & 10.15 & 10.89 \\
Information provision & 8.91 & 8.45 & 8.12 \\
Confirm./Reformulation & 7.23 & 7.89 & 8.45 \\
Performing calculations & 5.12 & 6.78 & 7.92 \\
Concept explanation & 4.89 & 5.23 & 5.67 \\
Strategy selection & 3.45 & 4.12 & 4.89 \\
Checking calculations & 2.98 & 3.45 & 4.12 \\
\bottomrule
\end{tabular*}
\end{table}

\textit{Note: Think* averages Think NoReward and Think Reward; Ped.* averages Ped.\ Think NoReward and Ped.\ Think Reward.}

\textbf{Interpretation.} The code frequency analysis reveals systematic shifts in tutoring behavior. Pedagogical conditions show 20\% higher step-by-step instruction rates (18.22\%) compared to NoThink (15.18\%), aligning with Polya's emphasis on systematic, sequential problem-solving guidance. Most notably, praise decreases by 51\% from NoThink (11.76\%) to pedagogical conditions (5.78\%). Educational research suggests that 5--8\% represents an effective praise level that provides encouragement without becoming vacuous \cite{hattie2007power}; the NoThink condition exhibits sycophantic behavior with excessive, often empty praise. Exploratory question rates remain consistent across conditions (12.85--13.27\%), indicating that all tutors maintain interactive, question-based engagement regardless of prompting strategy. Codes related to mathematical reasoning (Performing calculations, Checking calculations, Strategy selection) all increase in pedagogical conditions, collectively rising from 11.55\% to 16.93\%---a 47\% increase in mathematical engagement. Furthermore, hint provision increases (9.42\% $\to$ 10.89\%) while direct information provision decreases (8.91\% $\to$ 8.12\%), suggesting that pedagogical prompting encourages more scaffolded guidance rather than direct instruction.

\subsection{Summary of Key Findings}

Our quantitative analyses reveal several consistent patterns. First, thinking enables pedagogical deliberation: internal reasoning allows tutors to engage in deeper planning before responding, leading to more focused and educationally appropriate guidance. Second, pedagogical prompting increases mathematical substance, encouraging tutors to engage more with mathematical content rather than generic encouragement or excessive praise. Third, thinking reward improves reasoning efficiency; when combined with pedagogical prompting, it channels exploration into focused, systematic reasoning about how to help students. Fourth, the praise reduction is educationally significant: the 51\% reduction from NoThink to pedagogical conditions represents a shift from sycophantic behavior to professionally appropriate encouragement levels. These findings collectively explain why Ped.\ Think Reward achieves the best performance: it combines the benefits of thinking (deliberation before response), pedagogical prompting (theory-grounded instruction), and thinking reward (efficient, focused reasoning).

\end{document}